\documentclass[12pt,journal,compsoc,onecolumn]{IEEEtran}

\usepackage{amsmath,graphicx,cleveref,caption,subcaption,dsfont,multirow,multicol,amssymb,amsfonts,mathtools,pifont}
\usepackage{booktabs}
\usepackage{lipsum}
\usepackage{bm}
\usepackage{stmaryrd}
\usepackage{algpseudocode}
\usepackage[ruled,vlined]{algorithm2e}
\usepackage[noadjust]{cite}
\let\OLDthebibliography\thebibliography
\renewcommand\thebibliography[1]{
  \OLDthebibliography{#1}
  \setlength{\parskip}{0pt}
  \setlength{\itemsep}{1.47pt plus 0.3ex}
}

\newcommand{\cmark}{\ding{51}}
\newcommand{\xmark}{\ding{55}}

\def\U{{\mathbf U}} 
\def\V{{\mathbf V}} 
\def\A{{\mathbf \Lambda}}

\def\A{{\bf A}}

\def\U{{\bf U}}
\def\W{{\bf W}}
\def\WW{{\bf W}}
\def\SS{{\cal S}}
\def\S{{\bf S}}
\def\N{{\cal N}}
\def\G{{\cal G}}
\def\V{{\cal V}}
\def\E{{\cal E}}
\def\F{{\cal F}}

\def\M{{\bf M}}

\title{Lightweight Graph Convolutional Networks with Topologically Consistent Magnitude Pruning}
\author{Hichem Sahbi \\ Sorbonne University, UPMC, CNRS, LIP6, France}

\begin{document}
\maketitle
\begin{abstract}
Graph convolution networks (GCNs) are currently mainstream in learning with irregular data. These models rely on message passing and attention mechanisms that capture context and node-to-node relationships. With multi-head attention, GCNs become highly accurate but oversized, and their deployment on cheap devices requires their pruning. However, pruning at high regimes usually leads to topologically inconsistent networks with weak generalization.\\
In this paper, we devise a novel method for lightweight GCN design. Our proposed approach parses and selects subnetworks with the highest magnitudes while guaranteeing their topological consistency. The latter is obtained by selecting only accessible and co-accessible connections which actually contribute in the evaluation of the selected subnetworks. Experiments conducted on the challenging FPHA dataset show the substantial gain of our topologically consistent pruning method especially at very high pruning regimes.\\

\end{abstract}

  \section{Introduction}
  \label{sec:intro}
  
  Deep convolutional networks are currently one of the most successful models in  image processing and pattern recognition. Their principle consists in learning convolutional  filters, together with attention and fully connected layers, that maximize classification performances. These models are mainly suitable for data sitting on top of regular domains (such as images) \cite{Krizhevsky2012},  but their adaptation to irregular data (namely graphs) requires extending  convolutions to arbitrary domains \cite{Bruna2013,Henaff2015}; these extensions are known as graph convolutional networks (GCNs). \\
\indent Two categories of GCNs exist in the literature: spectral and spatial. Spectral methods \cite{kipf17,Levie2018,Li2018,Bresson16,sahbi00006,sahbi00004} proceed by projecting both input graph signals and convolutional filters using the Fourier transform, and achieve convolution in the Fourier domain, prior to back-project the resulting convolved signal in the input domain.  These projections rely on the eigen-decomposition of graph Laplacians whose complexity scales polynomially  with the size of the input graphs \cite{spectral1997}, and this makes spectral GCNs clearly intractable.  Spatial methods  \cite{Gori2005,Micheli2009,Wu2019,Hamilton2017,sahbi00005,sahbi00001} instead rely on message passing, via attention matrices,  before applying convolution.  While spatial GCNs have been relatively more effective compared to spectral ones, their success is highly reliant on the accuracy of the attention matrices that capture context and node-to-node relationships \cite{attention2019}.  With multi-head attention,  GCNs are more accurate but computationally more demanding, so lightweight variants of these models should instead be considered. \\
\indent Several methods have been proposed in the literature in order to design lightweight yet effective deep convolutional networks \cite{DBLP:conf/cvpr/HuangLMW18,DBLP:conf/cvpr/SandlerHZZC18,DBLP:journals/corr/HowardZCKWWAA17,DBLP:conf/icml/TanL19,icassp2017b,sahbiiccv17,sahbi00003}. Some of them build efficient networks from scratch while others pretrain heavy networks prior to reduce their time and memory footprint using distillation \cite{DBLP:journals/corr/HintonVD15,DBLP:conf/iclr/ZagoruykoK17,DBLP:journals/corr/RomeroBKCGB14,DBLP:conf/aaai/MirzadehFLLMG20,DBLP:conf/cvpr/ZhangXHL18,DBLP:conf/cvpr/AhnHDLD19} and pruning \cite{DBLP:conf/nips/CunDS89,DBLP:conf/nips/HassibiS92,DBLP:conf/nips/HanPTD15}.  Pruning methods, either unstructured or structured, allow removing connections whose impact on the classification performance is the least perceptible. Unstructured pruning \cite{DBLP:conf/nips/HanPTD15,DBLP:journals/corr/HanMD15} consists in cutting connections individually using different criteria, including weight magnitude\footnote{Magnitude is considered as a proxy of weight relevance.}, prior to fine-tuning.  In contrast,  structured pruning \cite{DBLP:conf/iclr/0022KDSG17,DBLP:conf/iccv/LiuLSHYZ17} aims at removing  groups of connections, channels or entire subnetworks. Whereas structured pruning may reach high speed-up on dedicated hardware resources, its downside resides in the rigidity of the class of learnable lightweight networks. On another side, unstructured pruning is more flexible, but may result into {\it topologically inconsistent} subnetworks (i.e., either partially or completely disconnected), and this may lead to limited generalization especially at very high pruning rates.   \\
\indent In this paper, we introduce a novel approach for lightweight GCN design that gathers the advantage of both structured and unstructured pruning, and discards their inconvenient; i.e., the method imposes a few constraints on the structure of the learned subnetworks (namely their topological consistency) while also ensuring their flexibility at some extent.  Our proposed solution is greedy and proceeds by selecting connections with the highest magnitudes while guaranteeing their {\it accessibility} (i.e., their reachability from the network input)  and  their {\it co-accessibility} (i.e., their actual contribution in the evaluation of the output). Hence,  only topologically consistent subnetworks are considered when selecting connections.  Different magnitude surrogates are also considered in order to greedily select connections; these surrogates allow maximizing magnitudes not only locally at the visited layers but also globally at the subsequent ones, thereby resulting into more effective lightweight networks as shown  in experiments.    

\section{Graph Convolutional Networks}
Let $\SS=\{\G_i=(\V_i, \E_i)\}_i$ denote a collection of graphs with $\V_i$, $\E_i$ being respectively the nodes and the edges of $\G_i$. Each graph $\G_i$ (denoted for short as $\G=(\V, \E)$) is endowed with a signal $\{\psi(u) \in \mathbb{R}^s: \ u \in \V\}$ and associated with an adjacency matrix $\A$ with each entry  $\A_{uu'}>0$ iff $(u,u') \in \E$ and $0$ otherwise. GCNs aim at learning a set of $C$ filters $\F$ that define convolution on $n$ nodes of $\G$ (with $n=|\V|$) as
 \begin{equation}\label{matrixform} 
\small  (\G \star \F)_\V = f\big(\A \  \U^\top  \   \W\big), 
 \end{equation} 
 \noindent  here $^\top$ stands for transpose,  $\U \in \mathbb{R}^{s\times n}$  is the  graph signal, $\W \in \mathbb{R}^{s \times C}$  is the matrix of convolutional parameters corresponding to the $C$ filters and  $f(.)$ is a nonlinear activation applied entrywise. In Eq.~\ref{matrixform}, the input signal $\U$ is projected using $\A$ and this provides for each node $u$, the  aggregate set of its neighbors. Entries of $\A$ could be handcrafted or learned so Eq.~(\ref{matrixform}) implements a convolutional block with two layers; the first one aggregates signals in $\N(\V)$ (sets of node neighbors) by multiplying $\U$ with $\A$ while the second layer achieves convolution by multiplying the resulting aggregates with the $C$ filters in $\W$. Learning  multiple adjacency (also referred to as attention) matrices (denoted as $\{\A^k\}_{k=1}^K$) allows us to capture different contexts and graph topologies when achieving aggregation and convolution.  With multiple matrices $\{\A^k\}_k$ (and associated convolutional filter parameters $\{\W^k\}_k$),  Eq.~\ref{matrixform} is updated as  $\small (\G \star \F)_\V = f\big(\sum_{k=1}^K \A^k   \U^\top     \W^k\big)$. Stacking aggregation and convolutional layers, with multiple  matrices $\{\A^k\}_k$, makes GCNs accurate but heavy. In what follows, we propose a novel method that makes our networks lightweight and still effective.

\section{Lightweight  Design}
In what follows, we subsume any given GCN as a multi-layered neural network $g_\theta$ whose weights defined as $\theta =  \left\{\WW^1,\dots, \WW^L \right\}$, with $L$ being its depth,  $\WW^\ell \in \mathbb{R}^{d_{\ell-1} \times d_{\ell}}$ its $\ell^\textrm{th}$  layer weights, and $d_\ell$ the dimension of $\ell$. The output of a given layer  $\ell$ is defined as 
  \begin{equation}
    \label{eq:layer_eq}
    \mathbf{\phi}^{\ell} = f_\ell({\WW^\ell}^\top \  \mathbf{\phi}^{\ell-1}), \ \ \ \ell \in \{1,\dots,L-1\}, 
  \end{equation}
being $f_\ell$ an activation function.  Without a loss of generality, we omit the bias in the definition of  (\ref{eq:layer_eq}).
\subsection{Magnitude Pruning}

Given a GCN $g_\theta$, magnitude pruning (MP) consists in removing connections in  $g_\theta$. MP is obtained by zeroing-out a subset of weights in $\theta$, and this  is achieved  by multiplying $\WW^\ell$ by a binary mask $\M^\ell \in \{ 0,1 \}^{d_{\ell-1} \times d_{\ell}}$. The binary entries of  $\M^\ell$ are set depending on whether the underlying layer connections are kept or removed, so Eq.~\ref{eq:layer_eq} becomes
  \begin{equation}
    \label{eq:pruned_layer_eq}
    \mathbf{\phi}^{\ell} = f_\ell( (\M^\ell \odot \WW^\ell )^\top \ \mathbf{\phi}^{\ell-1} ),
  \end{equation}
  here $\odot$ stands for the element-wise matrix product. In this definition, entries of the tensor $\{\M^\ell\}_\ell$ are set depending on the prominence of the underlying connections in $g_\theta$; MP consists first in zeroing the smallest parameters (up to a pruning rate) in the learned GCN $g_\theta$, and then fine-tuning the remaining parameters. However, such MP suffers from several drawbacks. On the one hand, removing connections individually may result into {\it topologically inconsistent} networks (see section~\ref{tc}), i.e., either completely disconnected or having isolated connections. On the other hand, high pruning rates may lead to an over-regularization effect and hence weakly discriminant lightweight networks, especially when the latter include isolated connections (see later experiments). In what follows, we introduce a more principled MP that guarantees the topological consistency of the pruned networks and allows improving generalization even at very high pruning rates. 
\subsection{Our Topologically Consistent Magnitude Pruning}\label{tc}
Our formal definition of topological consistency relies on two principles: {\it accessibility and co-accessibility} of connections in $g_\theta$. Let $\M_{i,j}^\ell$ refer to a connection between the i-th and the j-th neurons of layer $\ell$. $\M_{i,j}^\ell$ is accessible if $ \ \exists i_1,\dots,i_{\ell-1}$, s.t. $\M_{i_1,i_2}^1=\dots =\M_{i_{\ell-1},i}^{\ell-1}=1$, and  $\M_{i,j}^\ell$ is co-accessible if $\exists i_{\ell+1},\dots,i_L$, s.t. $\M_{j,i_{\ell+1}}^{\ell+1}= \dots =\M_{i_{L-1},i_L}^{L}=1$.  

Considering the products $\S_a^\ell = \M^1 \ \M^2 \dots \M^{\ell-1}$ and $\S_c^\ell = \M^{\ell+1} \ \M^{\ell+2} \dots \M^{L}$, and following the above definition, it is easy to see that $\M_{ij}^\ell$ is accessible (resp. co-accessible) iff the i-th column (resp. j-th row) of $\S_a^\ell$ (resp. $\S_c^\ell$) is different from the null vector. A network is called {topologically consistent} iff all its connections are both accessible and co-accessible. Accessibility guarantees that incoming connections to the i-th neuron carry out effective activations resulting from the evaluation of $g_\theta$ up to layer $\ell$. Co-accessibility is equivalently important and guarantees that outgoing activation from the j-th neuron actually contributes in the evaluation of the network output. A connection $\M_{ij}^\ell$ --- not satisfying accessibility or co-accessibility and even when its magnitude is large --- becomes useless and should be removed when $g_\theta$ is pruned.\\  \indent For any given network, parsing all its topologically consistent subnetworks and keeping only the one with the largest magnitudes is highly combinatorial. Indeed, the accessibility of a given connection depends on whether its  preceding and subsequent ones are kept or removed, and any masked connections may affect the accessibility of the others. In what follows, we introduce a greedy algorithm that prunes a given network by maximizing the magnitude of its connections while guaranteeing its topological consistency. 
\def\S{{\bf S}}
\begin{algorithm}[!ht]
\KwIn{Weight tensor $\left\{\WW^1,\dots, \WW^L \right\}$,  {MaxKeptConnections}.}
\KwOut{Mask tensor $\left\{\M^1,\dots, \M^L \right\}$.}
\BlankLine
${nc}  \leftarrow 0$;  $\{\M^\ell \leftarrow 0 \}_\ell$;\\
\While{${nc} < \textrm{MaxKeptConnections}$}{
Select $i_1$ from $\{1,\dots,d_1\}$; \\
\For{$\ell=1$ {\bf to} $L-1$}{ 
$i_{\ell+1} \leftarrow \arg\max_ {j\in {\cal N}_\ell(i_\ell),k} \  [ \WW_{i_\ell,j}^\ell \ \hat{\WW}_{j,k}^{\ell+1}]$  \tcp*[r]{A stochastic variant is to select a random walk from $i_\ell$ to $i_{\ell+1}  \in  {\cal N}_\ell(i_\ell)$ proportionally to $\{ \WW_{i_\ell,j}^\ell  \ \hat{\WW}_{j,k}^{\ell+1}\}_{j\in {\cal N}_\ell(i_\ell),k}$.} 
\If{($\M_{i_\ell,i_{\ell+1}}^\ell=0$)}{${nc} \leftarrow {nc} + 1$;} 
$\M_{i_\ell,i_{\ell+1}}^\ell \leftarrow 1$;
}
}
\caption{Topologically consistent MP}\label{alg1}
\end{algorithm}
\subsection{Algorithm} 
\indent Our solution parses neurons in $g_\theta$ layer-wise; each parsing consists in finding a complete chain from the input to the output of $g_\theta$. Given a neuron $i$ in layer $\ell$, its subsequent neuron in the chain corresponds to the one which maximizes magnitude among the forward neighbors of $i$ (denoted as ${\cal N}_\ell(i)$). This process is repeated for different input neurons till exhausting a targeted pruning rate. This solution maximizes magnitude locally; however, there is not guarantee that neurons visited in the subsequent layers will have sufficiently large  magnitude connections. In order to circumvent this issue, instead of locally maximizing $\{\WW_{i,j}^\ell\}_{j\in {\cal N}_\ell(i)}$, we {\it globally} maximize a surrogate criterion as   
\begin{equation}\label{eq0}
  \max_{j\in {\cal N}_\ell(i),k} \WW_{i,j}^\ell \ \hat{\WW}_{j,k}^{\ell+1}, \ \ \ k \in \{1,\dots,d_L\}, 
\end{equation} 
\noindent with $\hat{\WW}^{\ell+1} = {\WW}^{\ell+1} \dots {\WW}^{L}$ (see also algorithm~\ref{alg1}). Under row-stochasticity of $\{\WW^\ell\}_\ell$, the matrix $\hat{\WW}^{\ell+1}$ models an $m$-step markovian process (with $m=L-\ell$) where the conditional transition likelihood, between two neurons, is proportional to the sum of the conditional likelihoods of all the possible $m-1$ steps linking these two neurons. Nevertheless, Eq.~\ref{eq0} could be contaminated by a large number of small magnitude connections. This limitation motivates the introduction of a slight variant (called $\alpha$-powered magnitude) with $\hat{\WW}^{\ell+1}$ recursively defined as 
\begin{equation}\label{eq1}
\hat{\WW}^{\ell+1} = \bigg[[\WW^{\ell+1}]^{\frac{1}{\alpha}}  \  [\hat{\WW}^{\ell+2}]^{\frac{1}{\alpha}}\bigg]^{\alpha},  \ \ \  1/\alpha \in [1,+\infty[,
\end{equation} 
\noindent here the power is applied entrywise. When $\alpha \rightarrow 0$,  Eq.~\ref{eq0} captures the {\it largest magnitude path}\footnote{The magnitude of a path is defined as the sum of all its connection magnitudes.} outgoing from the i-th to the k-th output neuron  of $g_\theta$ (via j).  When $\alpha \in ]0,1[$, Eq.~\ref{eq0} models instead {\it the average of dominating magnitude paths} outgoing from the i-th neuron (again via j), so the effect of spurious (small magnitude) connections could be attenuated. 
\subsection{Stochasticity} 
In spite of making the selected subnetworks topologically consistent and hence effective  (as shown later in experiments), the aforementioned procedure is deterministic and relies on the hypothesis that only connections with the highest magnitudes are essential while in practice, other connections could also be used in order to {\it explore} further subnetworks. Hence, instead of considering a deterministic  parsing approach, we consider instead a stochastic sampling process. More precisely, neurons are again visited layer-wise, but the subsequent layer neurons are selected by sampling a random walk distribution (see the commented variant in algorithm~\ref{alg1}); note that neurons that maximize magnitude are still preferred (with a high probability), nevertheless other neurons will also be selected depending on their magnitude distribution. This stochastic variant turns out to be more effective, especially at high pruning regimes, as shown subsequently. 
 \section{Experiments}\label{sec:experiments}
 We evaluate the performance of our GCNs on the task of action recognition using  the First-Person Hand Action (FPHA) dataset~\cite{garcia2018}. The latter includes 1175 skeletons belonging to 45 action categories (with style, speed, scale and viewpoint variations). Each video (sequence of skeletons) is initially described with a  graph $\G = (\V,\E)$ where each node $v_j \in \V$ corresponds to the $j$-th hand-joint trajectory (denoted as $\{\hat{p}_j^t\}_t$)  and an edge $(v_j, v_i) \in  \E$ exists iff the $j$-th and the $i$-th trajectories are spatially connected. Each trajectory in $\G$ is processed using {\it temporal chunking}: first, the total duration of a  sequence is split into $M$ equally-sized temporal chunks ($M=32$ in practice), then the trajectory  coordinates  $\{\hat{p}_j^t\}_t$  are assigned to the $M$ chunks (depending on their time stamps) prior to concatenate the averages of these chunks. This produces the raw description (signal) of $v_j$. \\

 \noindent {\bf Implementation details and baseline GCN.} We trained the GCNs end-to-end using the Adam optimizer \cite{Adam2014} for 2,700 epochs  with a batch size equal to $600$, a momentum of $0.9$ and a global learning rate (denoted as $\nu(t)$)  inversely proportional to the speed of change of the cross entropy loss used to train our networks. When this speed increases (resp. decreases),   $\nu(t)$  decreases as $\nu(t) \leftarrow \nu(t-1) \times 0.99$ (resp. increases as $\nu(t) \leftarrow \nu(t-1) \slash 0.99$). In all these experiments, we use a GeForce GTX 1070 GPU (with 8 GB memory). We evaluate the performances using the 1:1 setting proposed in~\cite{garcia2018} with 600 action sequences for training and 575 for testing, and we report the average accuracy over all the classes of actions. The architecture of our baseline GCN (taken from \cite{sahbiICPR2020}) includes an attention layer of 16 heads applied to skeleton graphs whose nodes are encoded with 32-channels, followed by a convolutional layer of 128 filters,  and a dense fully connected layer. In total, this initial network is relatively heavy (for a GCN) and its number of parameters reaches 2 millions. Nevertheless, this GCN is accurate compared to the related work on the FPHA benchmark as shown in table~\ref{compare2}. Considering this GCN baseline, our goal is to make it lightweight while maintaining its high accuracy.\\ 
\begin{table}[ht]
\begin{center}
\resizebox{0.74\columnwidth}{!}{
\begin{tabular}{ccccc}
{\bf Method} & {\bf Color} & {\bf Depth} & {\bf Pose} & {$ \ \ \ \ \  \ \ \ \ \ \  \  $ \bf Accuracy (\%) $ \ \ \ \  \ \ \ \ \ \ \  \  $}\\
\hline
  Two stream-color \cite{refref10}   & \cmark  &  \xmark  & \xmark  &  61.56 \\
Two stream-flow \cite{refref10}     & \cmark  &  \xmark  & \xmark  &  69.91 \\ 
Two stream-all \cite{refref10}      & \cmark  & \xmark   & \xmark  &  75.30 \\
\hline 
HOG2-depth \cite{refref39}        & \xmark  & \cmark   & \xmark  &  59.83 \\    
HOG2-depth+pose \cite{refref39}   & \xmark  & \cmark   & \cmark  &  66.78 \\ 
HON4D \cite{refref40}               & \xmark  & \cmark   & \xmark  &  70.61 \\ 
Novel View \cite{refref41}          & \xmark  & \cmark   & \xmark  &  69.21  \\ 
\hline
1-layer LSTM \cite{Zhua2016}        & \xmark  & \xmark   & \cmark  &  78.73 \\
2-layer LSTM \cite{Zhua2016}        & \xmark  & \xmark   & \cmark  &  80.14 \\ 
\hline 
Moving Pose \cite{refref59}         & \xmark  & \xmark   & \cmark  &  56.34 \\ 
Lie Group \cite{Vemulapalli2014}    & \xmark  & \xmark   & \cmark  &  82.69 \\ 
HBRNN \cite{Du2015}                & \xmark  & \xmark   & \cmark  &  77.40 \\ 
Gram Matrix \cite{refref61}         & \xmark  & \xmark   & \cmark  &  85.39 \\ 
TF    \cite{refref11}               & \xmark  & \xmark   & \cmark  &  80.69 \\  
\hline 
JOULE-color \cite{refref18}         & \cmark  & \xmark   & \xmark  &  66.78 \\ 
JOULE-depth \cite{refref18}         & \xmark  & \cmark   & \xmark  &  60.17 \\ 
JOULE-pose \cite{refref18}         & \xmark  & \xmark   & \cmark  &  74.60 \\ 
JOULE-all \cite{refref18}           & \cmark  & \cmark   & \cmark  &  78.78 \\ 
\hline 
Huang et al. \cite{Huangcc2017}     & \xmark  & \xmark   & \cmark  &  84.35 \\ 
Huang et al. \cite{ref23}           & \xmark  & \xmark   & \cmark  &  77.57 \\  
  \hline
Our  GCN baseline                  & \xmark  & \xmark   & \cmark  & \bf86.08                                                
\end{tabular}}
 \end{center} 
\caption{Comparison of our baseline GCN against related work on FPHA.}\label{compare2}
\end{table}

 \noindent {\bf Lightweight CGN performances.}  Table.~\ref{table21} shows the accuracy of our lightweight GCNs for different pruning rates, and other settings including topological consistency (TC) and stochasticity.  From these results, the impact of TC is substantial on highly pruned GCNs. We also observe the positive impact of stochasticity which allows exploring different subnetworks. Note that the impact of TC is less important (and sometimes worse) with low pruning regimes; indeed, low pruning rates produce subnetworks with already enough (a large number of) connections and {\it having some of them neither accessible nor co-accessible} produces a well known regularization effect~\cite{dropconnect2013}. On another side, over-pruning networks, without TC, produces an over-regularization effect (i.e., under-fitting); the resulting lightweight networks become highly disconnected.  In contrast, TC ensures connectivity (accessibility and co-accessibility) in spite of learning very lightweight networks, it also mitigates under-fitting and thereby improves generalization (see again accuracy in table~\ref{table21} and how the A-C difference between standard MP and TC MP is accentuated as pruning rates increase). Finally, table~\ref{table21222} shows the impact of $\alpha$ (in Eq.~\ref{eq1}) on the performance of our lightweight GCNs.  From this table,  sufficiently (but not very) large $\alpha$ makes accuracy improving; as  $m$-step magnitude estimation takes into account the dominating magnitude paths, it is more robust.
\begin{table}[ht]
 \begin{center}
\resizebox{0.79\columnwidth}{!}{
  \begin{tabular}{ccccccc}    
   \rotatebox{30}{Pruning rates}  & \rotatebox{30}{TC} & \rotatebox{30}{Stochastic} &     \rotatebox{30}{\# parameters} & \rotatebox{30}{\% of A-C } &  \rotatebox{30}{Accuracy (\%)}   & \rotatebox{30}{Observation}  \\
 \hline
  \hline
        0 \%                                   &  NA        &   NA       &     1967616         &  100   &   86.08  & Baseline GCN        \\
        \hline
  \multirow{4}{*}{\rotatebox{0}{50\%}} &     \xmark &  \xmark  &    \multirow{4}{*}{\rotatebox{0}{983808}}      & 100 & 86.08   & Standard MP   \\
                                         &   \xmark        &   \cmark         &              &  100    &   86.08 & Stochastic MP        \\
                                         &      \cmark       &    \xmark         &              &  100    &   86.08  & TC MP       \\
                                         &       \cmark       &     \cmark         &              &  100    &86.08    & TC Stoch MP      \\
  \hline
  \multirow{4}{*}{\rotatebox{0}{75\%}} &     \xmark &  \xmark  &    \multirow{4}{*}{\rotatebox{0}{491904}}      & 99.4 & 85.73  & Standard MP     \\
                                         &   \xmark        &   \cmark         &              &  99.8    &   \bf85.91   & Stochastic MP     \\
                                         &      \cmark       &    \xmark         &              &  100    &   84.86    & TC MP    \\
                                         &       \cmark       &     \cmark         &              &  100    &   \bf85.91  & TC Stoch MP       \\
  \hline 
  \multirow{4}{*}{\rotatebox{0}{90\%}} &     \xmark &  \xmark  &    \multirow{4}{*}{\rotatebox{0}{196760}}     & 89.9   & 85.04   & Standard MP    \\
                                         &   \xmark        &   \cmark         &              &  92.3    &   \bf85.56 & Stochastic MP       \\
                                         &      \cmark       &    \xmark         &              &  100    &   83.65   &  TC MP      \\
                                         &       \cmark       &     \cmark         &              &  100    &   \bf85.56  & TC Stoch MP       \\
 \hline  \multirow{4}{*}{\rotatebox{0}{95\%}} &     \xmark &  \xmark  &    \multirow{4}{*}{\rotatebox{0}{98379}}      & 72.3  & 83.82    & Standard MP  \\
                                         &   \xmark        &   \cmark         &              &  76.5    &   85.39      & Stochastic MP   \\
                                         &      \cmark       &    \xmark         &              &  100    &   \bf85.73  &  TC MP     \\
                                         &       \cmark       &     \cmark         &              &  100    &   84.86  &  TC Stoch MP       \\
  \hline  \multirow{4}{*}{\rotatebox{0}{99\%}} &     \xmark &  \xmark  &    \multirow{4}{*}{\rotatebox{0}{19674}}     &  21.2   & 76.00  & Standard MP    \\
                                         &   \xmark        &   \cmark         &              &  28.2   &   74.08   &Stochastic MP      \\
                                         &      \cmark       &    \xmark         &              &  100    &   \bf83.47&  TC MP     \\
                                         &       \cmark       &     \cmark         &              &  100    &   80.69  &  TC Stoch MP      \\
 \hline   \multirow{4}{*}{\rotatebox{0}{99.9\%}} &     \xmark &  \xmark  &   \multirow{4}{*}{\rotatebox{0}{1966}}       &  1.2    & 2.78   & Standard MP    \\
                                         &   \xmark        &   \cmark         &           &  0.0    &  NA  & Disconnected    \\
                                         &      \cmark       &    \xmark         &             &  100    &   70.08   &  TC MP     \\
                                         &       \cmark       &     \cmark         &             &  100    &  \bf73.39   &  TC Stoch MP  
                                       
  \end{tabular}}
\end{center}
\caption{Detailed performances and ablation study, for different pruning rates (\# of parameters) and other criteria including topological consistency (TC) and stochasticity. This table also shows the resulting percentage of accessible and co-accessible connections (denoted as A-C). These results are obtained by maximizing magnitudes locally (without Eqs.~\ref{eq0} and \ref{eq1}, i.e., by maximizing $\{\WW_{i,j}^\ell\}_{j\in {\cal N}_\ell(i)}$ instead).  NA stands for not applicable.}\label{table21}
\end{table}
\begin{table}[ht]
 \begin{center}
\resizebox{0.77\columnwidth}{!}{
  \begin{tabular}{c|cccccccc}    
    ${1}\slash{\alpha}$  & 1 & 1.5  & 2.5 & 7 & 10 & 20& 50 \\
    \hline 
$ \ \ \ \ \ \ \ \ \ \ \ $   Accuracy (\%) $ \ \ \ \ \ \ \ \  \ \ \ $   &  69.56 &  71.82  &  72.34 & 73.21  & \bf76.17  & 72.86 &  66.60
  \end{tabular}}
\end{center}
\caption{Accuracy for different $\alpha$ settings when maximizing magnitudes globally (i.e., with Eqs.~\ref{eq0} and \ref{eq1}). Both TC and stochasticity are used, and pruning rate (PR) is set to 99.9 \%. Compared to table~\ref{table21}, with the same PR, the accuracy improves significantly when $\alpha$ is set appropriately ($\alpha=10$).}\label{table21222}
\end{table}
\section{Conclusion}
In this paper, we introduce a novel pruning method that trains very lightweight GCNs while guaranteeing their topological consistency. The latter is an important property which guarantees the contribution of all accessible and co-accessible network connections in the learned decision functions. Experiments conducted on the challenging task of hand-gesture recognition shows the maintained high accuracy of our topologically consistent lightweight GCNs, even at {\it very high} pruning regimes. As a future work, we are currently investigating the extension of this method to other network architectures and datasets. 

\vfill\pagebreak

{  

}

\end{document}